\documentclass[11pt,a4paper]{article}
\usepackage[hyperref]{acl2020}
\usepackage{times}
\usepackage{latexsym}

\usepackage{CJKutf8}
\usepackage{graphicx}
\usepackage{algorithm}
\usepackage{algorithmic}
\usepackage{float}
\usepackage[T2A,T1]{fontenc}
\usepackage[russian,english]{babel}
\usepackage{ amssymb }
\usepackage{booktabs}
\usepackage{footmisc}

\usepackage{url}

\usepackage{microtype}

\aclfinalcopy 
\def\aclpaperid{1546} 

\title{Bilingual Dictionary Based Neural Machine Translation without Using Parallel Sentences}

\author{
	{Xiangyu Duan\textsuperscript{1}, Baijun Ji\textsuperscript{1}, Hao Jia\textsuperscript{1}, Min Tan\textsuperscript{1}, Min Zhang\textsuperscript{1}\thanks{  $\quad $ Corresponding Author. }, } \\
	{\textbf{Boxing Chen\textsuperscript{2}, Weihua Luo\textsuperscript{2}, Yue Zhang\textsuperscript{3} } }
	\vspace{1.6mm}\\
	\fontsize{12}{10}\selectfont
	\,\textsuperscript{\rm 1}  Institute of Aritificial Intelligence, School of Computer Science and Technology, \\
	\fontsize{12}{10}\selectfont Soochow university  \\
            \fontsize{12}{10}\selectfont  \textsuperscript{\rm 2} Alibaba DAMO Academy \\
            \fontsize{12}{10}\selectfont  \textsuperscript{\rm 3} School of Engineering, Westlake University \\
            \fontsize{10}{10}\selectfont \{xiangyuduan,minzhang\}@suda.edu.cn;  \{bjji,hjia,mtan2017\}@stu.suda.edu.cn;\\
	\fontsize{10}{10}\selectfont \{boxing.cbx,weihua.luowh\}@alibaba-inc.com; yue.zhang@wias.org.cn\\	}

\date{}

\begin{document}
\maketitle
\begin{abstract}
In this paper, we propose a new task of machine translation (MT), which is based on no parallel sentences but can refer to a ground-truth bilingual dictionary. Motivated by the ability of a monolingual speaker learning to translate via looking up the bilingual dictionary, we propose the task to see how much potential an MT system can attain using the bilingual dictionary and large scale monolingual corpora, while is independent on parallel sentences. We propose anchored training (AT) to tackle the task. AT uses the bilingual dictionary to establish anchoring points for closing the gap between source language and target language. Experiments on various language pairs show that our approaches are significantly better than various baselines, including dictionary-based word-by-word translation, dictionary-supervised cross-lingual word embedding transformation, and unsupervised MT. On distant language pairs that are hard for unsupervised MT to perform well, AT performs remarkably better, achieving performances comparable to supervised SMT trained on more than 4M parallel sentences{\footnote {Code is available at \url{https://github.com/mttravel/Dictionary-based-MT} } }. 

\end{abstract}

\section{Introduction}

Motivated by a monolingual speaker acquiring translation ability by referring to a bilingual dictionary, we propose a novel MT task that no parallel sentences are available, while a ground-truth bilingual dictionary and large-scale monolingual corpora can be utilized. This task departs from unsupervised MT task that no parallel resources, including the ground-truth bilingual dictionary, are allowed to utilize ~\cite{Artetxe2018Unsupervised2,Lample2018Phrase}. This task is also distinct to supervised/semi-supervised MT task that mainly depends on parallel sentences ~\cite{Bahdanau2015Neural,Gehring2017Convolutional,Vaswani2017Attention,chen2018best,sennrich-haddow-birch2016Improving}. 

The bilingual dictionary is often utilized as a seed in bilingual lexicon induction (BLI) that aims to induce more word pairs within the language pair ~\cite{Mikolov2013Exploiting}. Another utilization of the bilingual dictionary is for translating low-frequency words in supervised NMT ~\cite{Arthur2016Incorporating,Zhang2016Bridging}. We are the first to utilize the bilingual dictionary and the large scale monolingual corpora to see how much potential an MT system can achieve without using parallel sentences. This is different from using artificial bilingual dictionaries generated by unsupervised BLI for initializing an unsupervised MT system ~\cite{Artetxe2018Unsupervised2,Artetxe2018Unsupervised,Lample2018Unsupervised}, we use the ground-truth bilingual dictionary and apply it throughout the training process.

We propose Anchored Training (AT) to tackle this task. Since word representations are learned over monolingual corpora without any parallel sentence supervision, the representation distances between source language and target language are often quite large, leading to significant translation difficulty. As one solution, AT selects words covered by the bilingual dictionary as anchoring points to drive the distance between the source language space and the target language space closer so that translation between the two languages becomes easier. Furthermore, we propose Bi-view AT that places anchors based on either source language view or target language view, and combines both views to enhance the translation quality. 

Experiments on various language pairs show that AT performs significantly better than various baselines, including word-by-word translation through looking up the dictionary, unsupervised MT, and dictionary-supervised cross-lingual word embedding transformation to make distances between both languages closer. Bi-view AT further improves AT performance due to mutual strengthening of both views of the monolingual data. When combined with cross-lingual pretraining ~\cite{Lample2019Cross}, Bi-view AT achieves performances comparable to traditional SMT systems trained on more than 4M parallel sentences. The main contributions of this paper are as follows:

\begin{itemize}

\item A novel MT task is proposed which can only use the ground-truth bilingual dictionary and monolingual corpora, while is independent on parallel sentences.

\item AT is proposed as a solution to the task. AT uses the bilingual dictionary to place anchors that can encourage monolingual spaces of both languages to become closer so that translation becomes easier.

\item  The detailed evaluation on various language pairs shows that AT, especially Bi-view AT, performs significantly better than various methods, including word-by-word translation, unsupervised MT, and cross-lingual embedding transformation. On distant language pairs that unsupervised MT struggled to be effective, AT and Bi-view AT perform remarkably better. 

\end{itemize}

\section{Related Work}

The bilingual dictionaries used in previous works are mainly for bilingual lexicon induction (BLI), which independently learns the embedding in each language using monolingual corpora, and then learns a transformation from one embedding space to another by minimizing squared euclidean distances between all word pairs in the dictionary ~\cite{Mikolov2013Exploiting,Artetxe2016Learning}. Later efforts for BLI include optimizing the transformation further through new training objectives, constraints, or normalizations ~\cite{Xing2015Normalized,Lazaridou2015Hubness,Zhang2016Ten,Artetxe2016Learning,Smith2017Offline,Faruqui2014Improving,Lu2015Deep}. Besides, the bilingual dictionary is also used for supervised NMT which requires large-scale parallel sentences ~\cite{Arthur2016Incorporating,Zhang2016Bridging}. To our knowledge, we are the first to use the bilingual dictionary for MT without using any parallel sentences. 

Our work is closely related to unsupervised NMT (UNMT) ~\cite{Artetxe2018Unsupervised2,Lample2018Phrase,yang-etal-2018-unsupervised,sun-etal-2019-unsupervised}, which does not use parallel sentences neither. The difference is that UNMT may use the artificial dictionary generated by unsupervised BLI for initialization ~\cite{Artetxe2018Unsupervised2,Lample2018Unsupervised} or abandon the artificial dictionary by using joint BPE so that multiple BPE units can be shared by both languages ~\cite{Lample2018Phrase}. We use the ground-truth dictionary instead and apply it throughout a novel training process. UNMT works well on close language pairs such as English-French, while performs remarkably bad on distant language pairs in which aligning the embeddings of both side languages is quite challenging. We use the ground-truth dictionary to alleviate such problem, and experiments on distant language pairs show the necessity of using the bilingual dictionary.

Other utilizations of the bilingual dictionary for tasks beyond MT include cross-lingual dependency parsing ~\cite{Xiao2014Distributed},  unsupervised cross-lingual part-of-speech tagging and semi-supervised cross-lingual super sense tagging ~\cite{Gouws2015Simple}, multilingual word embedding training ~\cite{Ammar2016Massively,Long2016Learning}, and transfer learning for low-resource language modeling ~\cite{Adams2017Cross}.

\section{Our Approach}

There are multiple freely available bilingual dictionaries such as Muse dictionary\footnote{https://github.com/facebookresearch/MUSE} ~\cite{conneau2017word}, Wiktionary\footnote{https://en.wiktionary.org/wiki/Wiktionary:Main\_Page}, and PanLex\footnote{https://panlex.org/}. We adopt Muse dictionary which contains 110 large-scale ground-truth bilingual dictionaries.

\begin{figure*}[htbp]
\flushleft
\centerline {\includegraphics[height=8cm,width=13cm]{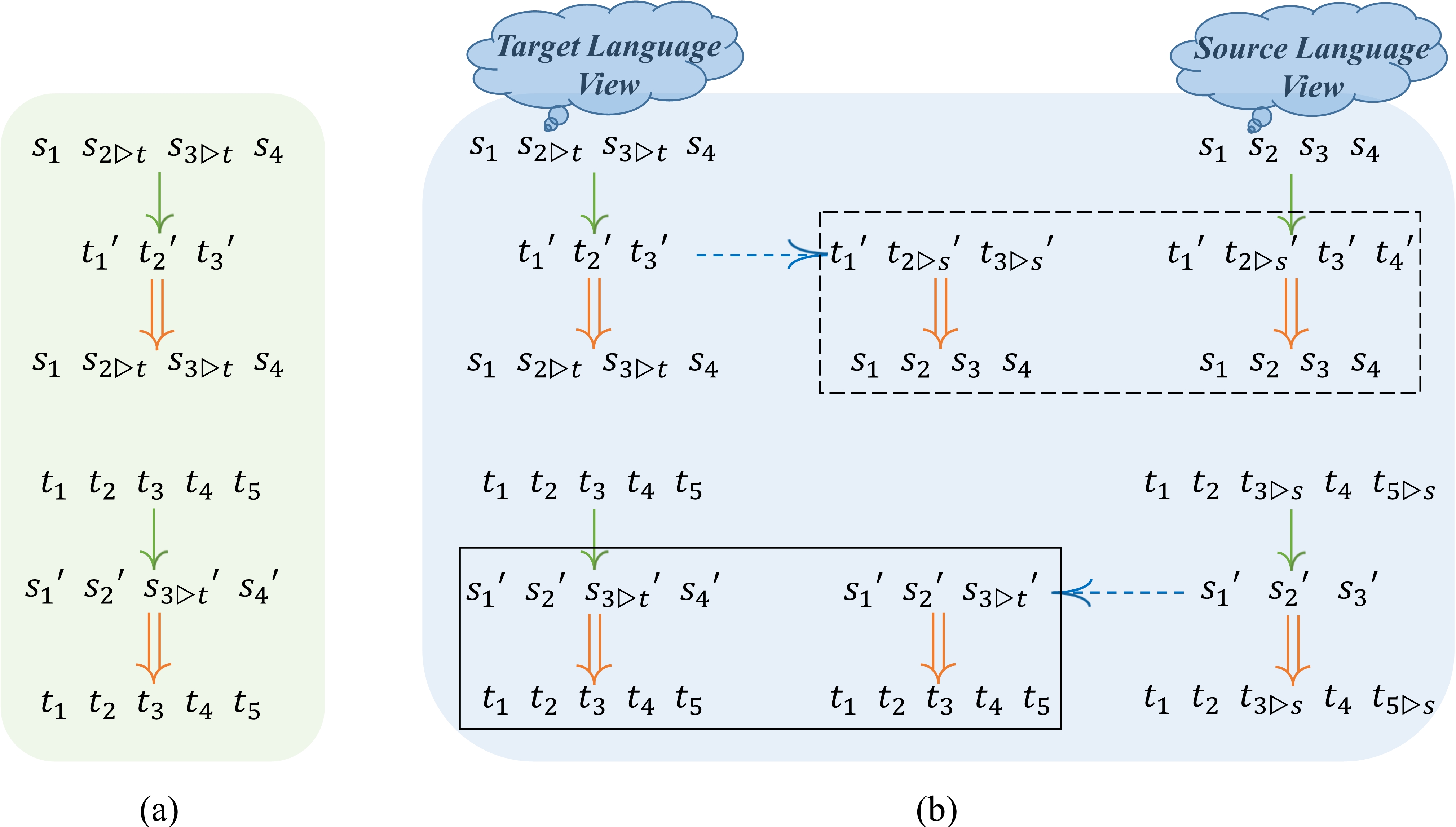} }
\caption{Illustration of (a) AT and (b) Bi-view AT. We use a source language sentence ``$s_1 s_2 s_3 s_4$'' and a target language sentence ``$t_1 t_2 t_3 t_4 t_5$'' from the large-scale monolingual corpora as an example. $\triangleright$ denotes an anchoring point which replaces a word with its translation based on the bilingual dictionary. Thin arrows of $\downarrow$ denote NMT decoding, thick arrows of $\Downarrow$ denote training an NMT model, $\dashrightarrow$ and $\dashleftarrow$ denote generating the anchored sentence based on the dictionary. Words with primes such as ${t_1}'$ denote the decoding output of a thin arrow.}
\label{fig:overall}
\end{figure*}

We propose to inject the bilingual dictionary into the MT training by placing anchoring points on the large scale monolingual corpora to drive the semantic spaces of both languages becoming closer so that MT training without parallel sentences becomes easier. We present the proposed Anchored Training (AT) and Bi-view AT in the following.

\subsection{Anchored Training (AT)}

Since word embeddings are trained on monolingual corpora independently, the embedding spaces of both languages are quite different, leading to significant translation difficulty. AT forces words of a translation pair to share the same word embedding as an anchor. We place multiple anchors by selecting words covered by the bilingual dictionary. With stable anchors, the embedding spaces of both languages become more and more close during the AT process. 

As illustrated in Figure \ref{fig:overall} (a), given the source sentence ``$s_1 s_2 s_3 s_4$'' with words of $s_2$ and $s_3$ being covered by the bilingual dictionary, we replace the two words with their translation words according to the dictionary. This results in the source sentence ``$s_1\ s_{2 \triangleright{t}}\ s_{3 \triangleright{t}}\ s_4$'', of which $s_{2 \triangleright{t}}$ and $s_{3 \triangleright{t}}$ serve as the \textbf{anchors} which are actually the target language words obtained by translating $s_2$ and $s_3$ according to the dictionary, respectively. Through the anchors, some words on the source side share the same word embeddings with the corresponding words on the target side. The AT process will strengthen the consistency of embedding spaces of both languages based on these anchors.

The training process illustrated in Figure \ref{fig:overall} (a) consists of a mutual back-translation procedure. The anchored source sentence ``$s_1\ s_{2 \triangleright{t}}\ s_{3 \triangleright{t}}\ s_4$'' is translated into target sentence ``${t_1}'\ {t_2}'\ {t_3}'$'' by using source-to-target \textbf{decoding}, then ``${t_1}'\ {t_2}'\ {t_3}'$'' and ``$s_1\ s_{2 \triangleright{t}}\ s_{3 \triangleright{t}}\ s_4$'' constitute a sentence pair for \textbf{training} the target-to-source translation model. In contrast, the target sentence ``$t_1 t_2 t_3 t_4 t_5$'' is translated into anchored source sentence ``${s_1}'\ {s_2}'\ {s_{3 \triangleright{t}}}'\ {s_4}'$'' by using target-to-source decoding, then both sentences constitute a sentence pair for training the source-to-target translation model. Note that during \textbf{training} the translation model, the input sentences are always pseudo sentences generated by decoding an MT model, while the output sentences are always true or anchored true sentences. Beside this mutual back-translation procedure, a denoising procedure used in unsupervised MT ~\cite{Lample2018Phrase} is also adopted. The deletion and permutation noises are added to the source/target sentence, and the translation model is also trained to denoise them into the original source/target sentence.

During testing, a source sentence is transformed into an anchored sentence at first by looking up the bilingual dictionary. Then we use the source-to-target model trained in the AT process to decode the anchored sentence.

We use Transformer architecture ~\cite{Vaswani2017Attention} as our translation model with four stacked layers in both encoder and decoder. In the encoder, we force the last three layers shared by both languages, and leave the first layer not shared. In the decoder, we force the first three layers shared by both languages, and leave the last layer not shared. Such architecture is designed to capture both common and specific characteristics of the two languages in one model for the training.

\subsection{Bi-view AT}

AT as illustrated in Figure \ref{fig:overall} (a) actually tries to model the sentences of both languages in the target language view with partial source words replaced with the target words and the full target language sentence. Bi-view AT enhances AT by adding another language view. Figure \ref{fig:overall} (b) adds the source language view shown in the right part to accompany with the target language view of Figure \ref{fig:overall} (a). In particular, the target language sentence ``$t_1 t_2 t_3 t_4 t_5$'' is in the form of ``$t_1\ t_2\ t_{3 \triangleright{s}}\ t_4\ t_{5 \triangleright{s}}$'' after looking up the bilingual dictionary. Such partial target words replaced with the source words and the full source language sentence ``$s_1 s_2 s_3 s_4$'' constitute the source language view. 

Based on the target language view shown in the left part and the source language view shown in the right part, we further combine both views through the pseudo sentences denoted by primes in Figure \ref{fig:overall} (b). As shown by ``$\dashrightarrow$'' in Figure \ref{fig:overall} (b), ``${t_1}' {t_2}' {t_3}'$'' is further transformed into ``${t_1}'\ {t_{2 \triangleright{s}}}'\ {t_{3 \triangleright{s}}}'$'' by looking up the bilingual dictionary. Similarly, ``${s_1}' {s_2}' {s_3}'$'' is further transformed into ``${s_1}' {s_2}' {s_{3 \triangleright{t}}}'$'' as shown by ``$\dashleftarrow$''. Finally, solid line box represents training the source-to-target model on data from both views, and dashed line box represents training the target-to-source model on data from both views.

Bi-view AT starts from training both views in parallel. After both views converge, we generate pseudo sentences in both the solid line box and the dashed line box, and pair these pseudo sentences (as input) with genuine sentences (as output) to train the corresponding translation model. This generation and training process iterates until Bi-view AT converges. Through such rich views, the translation models of both directions are mutually strengthened.

\subsection{Anchored Cross-lingual Pretraining (ACP)}

Cross-lingual pretraining has demonstrated effectiveness on tasks such as cross-lingual classification, unsupervised MT ~\cite{Lample2019Cross}. It is conducted over large monolingual corpora by masking random words and training to predict them as a cloze task. Instead, we propose ACP to pretrain on data that is obtained by transforming the genuine monolingual corpora of both languages into the anchored version. For example, words in the source language corpus that are covered by the bilingual dictionary are replaced with their translation words respectively. Such words are anchoring points that can drive the pretraining to close the gap between the source language space and the target language space better than the original pretraining method of Lample and Conneau \shortcite{Lample2019Cross} does as evidenced by the experiments in section \ref{pretrain-results}. Such anchored source language corpus and the genuine target language corpus constitute the target language view for ACP.

ACP can be conducted in either the source language view or the target language view. After ACP, each of them is used to initialize the encoder of the corresponding AT system. 

\subsection{Training Procedure}

For AT, the pseudo sentence generation step and NMT training step are interleaved. Take the target language view AT shown in Figure \ref{fig:overall} (a) for example, we extract anchored source sentences as one batch, and decode them into pseudo target sentences; then we use the same batch to train the NMT model of target-to-anchored source. In the meantime, a batch of target sentences are decoded into pseudo anchored source sentences, and then we use the same batch to train the NMT model of anchored source-to-target. The above process repeats until AT converges.

For Bi-view AT, after each mono-view AT converging, we set larger batch for generating pseudo sentences as shown in solid/dashed line boxes in Figure \ref{fig:overall} (b), and train the corresponding NMT model using the same batch.

For ACP, we follow XLM procedure ~\cite{Lample2019Cross}, and conduct pretraining on the anchored monolingual corpora concatenated with the genuine corpora of the other language.

\section{Experimentation}

We conduct experiments on English-French, English-Russian, and English-Chinese translation to check the potential of our MT system with only bilingual dictionary and large scale monolingual corpora. The English-French task deals with the translation between close-related languages, while the English-Russian and English-Chinese tasks deal with the translation between distant languages that do not share the same alphabets.

\subsection{Datasets}

For English-French translation task, we use the monolingual data released by XLM \cite{Lample2019Cross}\footnote{https://github.com/facebookresearch/XLM/blob/master/get-data-nmt.sh}. For English-Russian translation task, we use the monolingual data identical to  Lample et al.\shortcite{Lample2018Unsupervised}, which uses all available sentences for the WMT monolingual News Crawl datasets from years 2007 to 2017. For English-Chinese translation task, we extract Chinese sentences from half of the 4.4M parallel sentences from LDC, and extract English sentences from the complementary half. We use WMT \textit{newstest}-2013/2014, WMT \textit{newstest}-2015/2016, and NIST2006/NIST2002 as validation/test sets for English-French, English-Russian, and English-Chinese, respectively.

For cross-lingual pretraining, we extract raw sentences from Wikipedia dumps, which contain 80M, 60M, 13M, 5.5M monolingual sentences for English, French, Russian, and Chinese, respectively.


Muse ground-truth bilingual dictionaries are used for our dictionary-related experiments. If a word has multiple translations, we select the translation word that appears most frequently in the monolingual corpus. Table \ref{tbl:dict} summarizes the number of word pairs and their coverage on the monolingual corpora on the source side. 

\begin{table}[htb]
  \small
  \centering
  \begin{tabular}{l | l | l }
   & entry no. & coverage  \\
   \hline
   fr$\rightarrow$en &  97,046 & 60.91\% \\
   en$\rightarrow$fr &  94,719 & 69.77\% \\
   ru$\rightarrow$en & 45,065 & 65.43\% \\
   en$\rightarrow$ru & 42,725 & 88.77\% \\
   zh$\rightarrow$en & 13,749 & 50.20\% \\
   en$\rightarrow$zh & 32,495 & 47.02\% \\
  \end{tabular}
  \caption{Statistics of Muse bilingual dictionaries.}\label{tbl:dict}
\end{table}

\subsection{Experiment Settings}

For AT/Bi-view AT without cross-lingual pretraining, we use Transformer with 4 layers, 512 embedding/hidden units, and 2048 feed-forward filter size, for fair comparison to UNMT ~\cite{Lample2018Phrase}. For AT/Bi-view AT with ACP, we set Transformer with 6 layers, 1024 embedding/hidden units, and 4096 feed-forward filter size for a fair comparison to XLM ~\cite{Lample2019Cross}.

We conduct joint byte-pair encoding (BPE) on the monolingual corpora of both languages with a shared vocabulary of 60k tokens for both English-French and English-Russian tasks, and 40k tokens for English-Chinese task \cite{Sennrich2015Neural}.

During training, we set the batch size to 32 and limit the sentence length to 100 BPE tokens. We employ the Adam optimizer with $lr = 0.0001$, $t_{warm\_up} = 4000$ and $dropout = 0.1$. At decoding time, we generate greedily with length penalty $\alpha = 1.0$.

\subsection{Baselines}

\begin{itemize}

\item Word-by-word translation by looking up the ground truth dictionary or the artificial dictionary generated by Conneau et al. \shortcite{conneau2017word}.

\item Unsupervised NMT (UNMT) that does not rely on any parallel resources ~\cite{Lample2018Phrase}\footnote{https://github.com/facebookresearch/UnsupervisedMT}. Besides, cross-lingual pretraining (XLM) based UNMT ~\cite{Lample2019Cross}\footnote{https://github.com/facebookresearch/XLM}, is also set as a stronger baseline (XLM+UNMT).

\item We implement a UNMT initialized by Unsupervised Word Embedding Transformation (UNMT+UWET) as a baseline\cite{artetxe2018iclr}. The transformation function is learned in an unsupervised way without using any ground-truth bilingual dictionaries~\cite{conneau2017word}\footnote{https://github.com/facebookresearch/MUSE\label{fn:repeat}}.

\item We also implement a UNMT system initialized by Supervised Word Embedding Transformation (UNMT+SWET) as a baseline. Instead of UWET used in Artetxe et al. \shortcite{artetxe2018iclr}, we use the ground-truth bilingual dictionary as the supervision signal to train the transformation function for transforming the source word embeddings into the target language space~\cite{conneau2017word}. After such initialization, the gap between the embedding spaces of both languages is narrowed for easy UNMT training.

\end{itemize}

\begin{table*}[htbp]
\small
\centering
\begin{tabular}{l|c|c|c|c|c|c}
\bottomrule[1.2pt]
system & fr $\rightarrow$ en & en $\rightarrow$ fr & ru $\rightarrow$ en & en $\rightarrow$ ru & zh $\rightarrow$ en & en $\rightarrow$ zh \\ \hline \hline
\multicolumn{7}{c}{Without Cross-lingual Pre-training} \\ \hline 
Word-by-word using artificial dictionary &  7.76& 4.88 &3.05  & 1.60 & 1.99 &1.14  \\
 Word-by-word using ground-truth dictionary&  7.97& 6.61 &4.17  & 2.81 & 2.68 &1.79  \\
 UNMT ~\cite{Lample2018Phrase} & 24.02 & 25.10 & 9.09 & 7.98 & 1.50  & 0.45 \\
 UNMT+SWET & 21.11 & 21.22 & 9.79 & 4.07 &  19.78& 7.84  \\
 UNMT+UWET & 19.80 & 21.27 & 8.79 & 6.21 & 15.54 & 6.62 \\
 \hline
 AT & 25.07 & 26.36 & 10.20 & 9.91 & 19.83 & 9.18 \\
 Bi-view AT & \textbf{27.11} & \textbf{27.54} & \textbf{12.85} & \textbf{10.64} & \textbf{21.16} & \textbf{11.23} \\ 
\bottomrule[1.2pt]
\multicolumn{7}{c}{With Cross-lingual Pre-training} \\ \hline 
 XLM+UNMT ~\cite{Lample2019Cross} &33.28& 35.10 & 17.39 & 13.29 & 20.68 & 11.28  \\ 
 \hline
 ACP+AT & 33.51 & 36.15 & 16.41 & 15.43 &  26.80 & 13.91 \\
 ACP+Bi-view AT & \textbf{34.05} & \textbf{36.56} & \textbf{20.09} & \textbf{17.62} &  \textbf{30.12} & \textbf{17.05} \\
 \toprule[1.2pt]
 Supervised SMT & - & - & 21.48 & 14.54 &  31.86& 16.55 \\
  \toprule[1.2pt]
\end{tabular}
\caption{Experiment results evaluated by BLEU using the multi-bleu script. }
\label{tbl:mainResult}
\end{table*}

\subsection{Experimental Results: without Cross-lingual Pretraining}

The upper part of Table \ref{tbl:mainResult} presents the results of various baselines and our AT approaches. AT and Bi-view AT significantly outperform the baselines, and Bi-view AT is consistently better than AT. Detailed comparisons are listed as below:

\vspace{6 pt}
\noindent  \textbf{Results of Word-by-word Translation}
\vspace{6 pt}

\noindent It shows that using the ground-truth dictionary is slightly better than using the artificial one generated by Conneau et al. \shortcite{conneau2017word}. Both performances are remarkably bad, indicating that simple word-by-word translation is not qualified as an MT method. More effective utilization of the bilingual dictionary is needed to improve the translation performance.

\vspace{6 pt}
\noindent  \textbf{Comparison between UNMT and UNMT with WET Initialization}
\vspace{6 pt}

\noindent UNMT-related systems generally improves the performance of the word-by-word translation. On the close-related language pair of English-French, UNMT is better than UNMT+UWET/SWET. This is partly because there are numerous BPE units shared by both English and French, enabling easy establishing the shared word embedding space of both languages. In contrast, WET that transforms the source word embedding into the target language space seems not a necessary initialization step since shared BPE units already establish the shared space.

On distant language pairs, UNMT does not have an advantage over UNMT with WET initialization. Especially on English-Chinese, UNMT performs extremely bad, even worse than the word-by-word translation method. We argue that this is because the BPE units shared by both languages are so few that UNMT fails to align the language spaces. In contrast, using the bilingual dictionary greatly alleviate such problem for distant language pairs. UNMT+SWET, which transforms the source word embedding into the target word embedding space supervised by the bilingual dictionary, outperforms UNMT by more than 18 BLEU points on Chinese-to-English and more than 7 BLEU points on English-to-Chinese. This indicates the necessity of the bilingual dictionary for translation between distant language pairs.

\vspace{6 pt}
\noindent  \textbf{Comparison between AT/Bi-view AT and The Baselines}
\vspace{6 pt}

\noindent Our proposed AT approaches significantly outperform the baselines. The baselines of using the ground-truth bilingual dictionary, i.e., word-by-word translation using the dictionary and UNMT+SWET that uses the dictionary to supervise the word embedding transformation, are inferior to our AT approaches.

The AT approaches consistently improves the performances over both close-related language pair of English-French and distant language pairs of English-Russian and English-Chinese. Our Bi-view AT achieves the best performance on all language pairs.

\subsection{Experimental Results: with Cross-lingual Pretraining} \label{pretrain-results}



The bottom part of Table \ref{tbl:mainResult} reports performances of UNMT with XLM, which conducts the cross-lingual pretraining on concatenated non-parallel corpora ~\cite{Lample2019Cross}, and performances of our AT/Bi-view AT with the anchored cross-lingual pretraining, i.e., ACP. The results show that our proposed AT approaches are still superior when equipped with the cross-lingual pretraining. 

\begin{figure}[htbp]
\centering
\centerline {\includegraphics[scale=0.56]{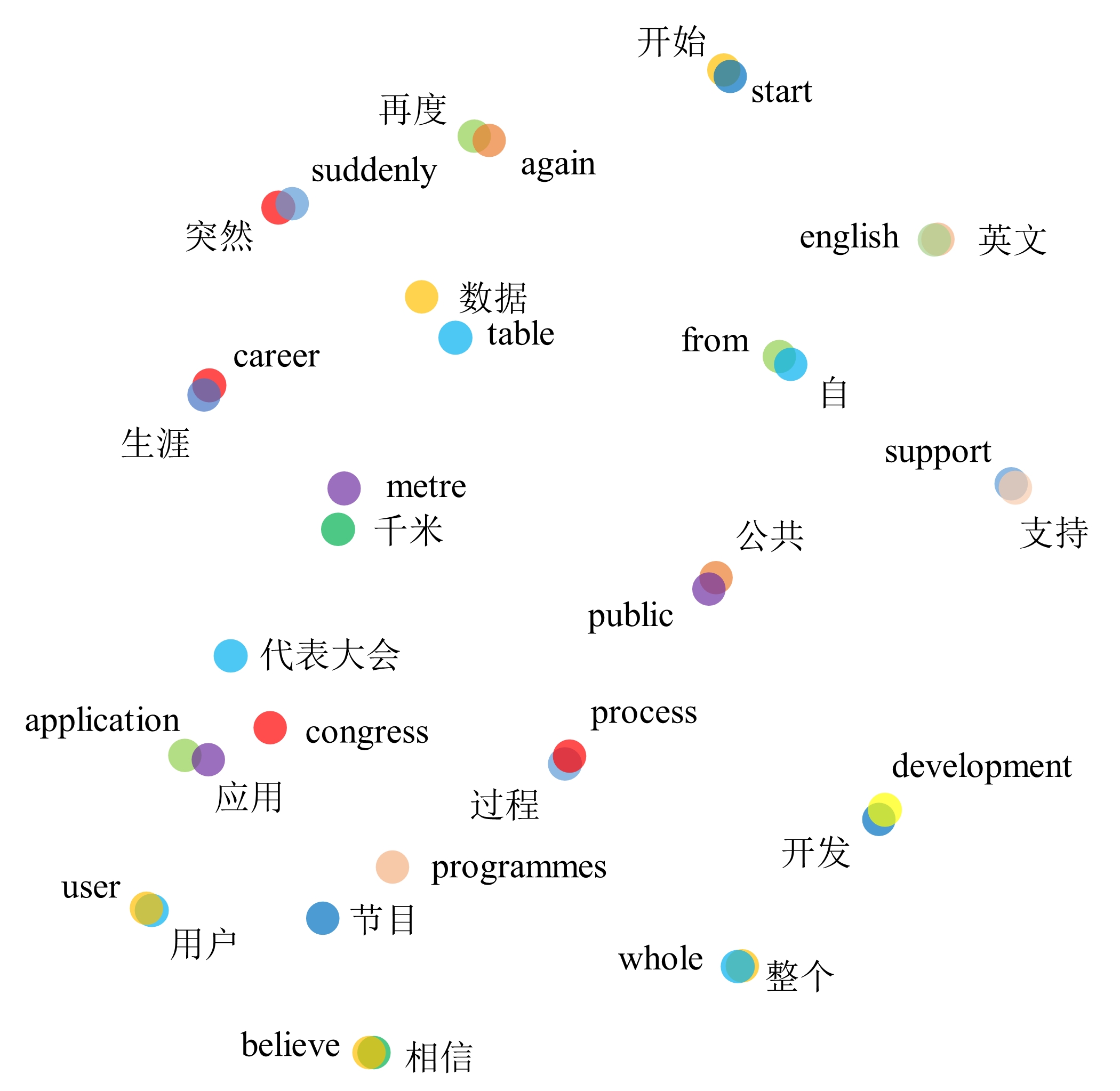} }
\caption{Visualization of the bilingual word embeddings after Bi-view AT.}
\label{fig:embedding}
\end{figure}

UNMT obtains great improvement when combined with XLM, achieving state-of-the-art unsupervised MT performance better than Unsupervised SMT ~\cite{artetxe2019acl-effective} and Unsupervised NMT ~\cite{Lample2018Phrase} across close and distant language pairs.

ACP+AT/Bi-view AT performs consistently superior to XLM+UNMT. Especially on distant language pairs, ACP+Bi-view AT gains 2.7-9.4 BLEU improvements over the strong XLM+UNMT. This indicates that AT/Bi-view AT with ACP builds closer language spaces via anchored pretraining and anchored training. We present such advantage in the analyses of Section \ref{sec:analysis}.

\vspace{6 pt}
\noindent  \textbf{Comparison with Supervised SMT}
\vspace{6 pt}

\noindent To check the ability of our system using only the dictionary and non-parallel corpora, we make the comparison to supervised SMT trained on over 4M parallel sentences, which are from WMT19 for English-Russian and from LDC for English-Chinese. We use Moses\footnote{http://www.statmt.org/moses/. We use the default setting of Moses.} as the supervised SMT system with a 5-gram language model trained on the target language part of the parallel corpora. 

The bottom part of Table \ref{tbl:mainResult} shows that ACP+Bi-view AT performs comparable to supervised SMT, and performs even better on English-to-Russian and English-to-Chinese.

\subsection{Analyses} \label{sec:analysis}

We analyze the cross-lingual property of our approaches in both word level and sentence level. We also compare the performances between the ground-truth dictionary and the artificial dictionary. In the end, we vary the size of the bilingual dictionary and report its impact on the AT training.

\vspace{5 pt}
\noindent  \textbf{Effect on Bilingual Word Embeddings}
\vspace{5 pt}

\noindent As shown in Figure \ref{fig:embedding}, we depict the word embeddings of some sampled words in English-Chinese after our Bi-view AT. The dimensions of the embedding vectors are reduced to two by using T-SNE and are visualized by the visualization tool in Tensorflow\footnote{https://projector.tensorflow.org/}.

We sample the English words that are not covered by the dictionary at first, then search their nearest Chinese neighbors in the embedding space. It shows that the words which constitute a new ground-truth translation pair do appear as neighboring points in the 2-dimensional visualization of Figure \ref{fig:embedding}.

\vspace{5 pt}
\noindent  \textbf{Precision of New Word Pairs}
\vspace{5 pt}

\begin{table*}[htb]
  \small
  \centering
  \begin{tabular}{c | c | c | c | c | c | c }
  \bottomrule[1.5pt]
  & MuseUnsupervised & MuseSupervised & UNMT+SWET & UNMT+UWET & AT & Bi-view AT \\
   \hline
   Precision@1 &  30.51 & 35.38 & \textbf{48.01} & 45.85 & 43.32 & 45.49 \\
   Precision@5 &  55.42 & 58.48 & 68.05  &  67.15 &  68.23  & \textbf{72.02} \\
   Precision@10 & 62.45 & 63.18 & 72.02  & 72.20 & 73.83  &  \textbf{76.71} \\
   \toprule[1.5pt]
  \end{tabular}
  \caption{Precision of Discovered New Word Pairs.}\label{tbl:precision}
\end{table*}

\noindent We go on with studying bilingual word embedding by quantitative analysis of the new word pairs, which are detected by searching bilingual words that are neighbors in the word embedding space, and evaluate them using the ground-truth bilingual dictionary. In particular, we split the Muse dictionary of Chinese-to-English into standard training set and test set as in BLI ~\cite{Artetxe2018Robust}. The training set is used for the dictionary-based systems, including our AT/Bi-view AT, UNMT+SWET, and Muse, which is a BLI toolkit. The test set is used to evaluate these systems by computing the precision of discovered translation words given the source words in the test set. The neighborhood is computed by CSLS distance ~\cite{conneau2017word}.

Table \ref{tbl:precision} shows the precision, where precision@k indicates the accuracy of top-k predicted candidate. Muse induces new word pairs through either the supervised way or the unsupervised way. MuseSupervised is better than MuseUnsupervised since it is supervised by the ground-truth bilingual dictionary. Our AT/Bi-view AT surpasses MuseSupervised by a large margin. UNMT+SWET/UWET also obtains good performance through the word embedding transformation. Bi-view AT significantly surpasses UNMT+SWET/UWET in precision@5 and precision@10, while is worse than them in precision@1. This indicates that Bi-view AT can produce better $n$-best translation words that are beneficial for NMT beam decoding to find better translations.

\begin{figure}[htbp]
\centering
\centerline {\includegraphics[scale=0.45]{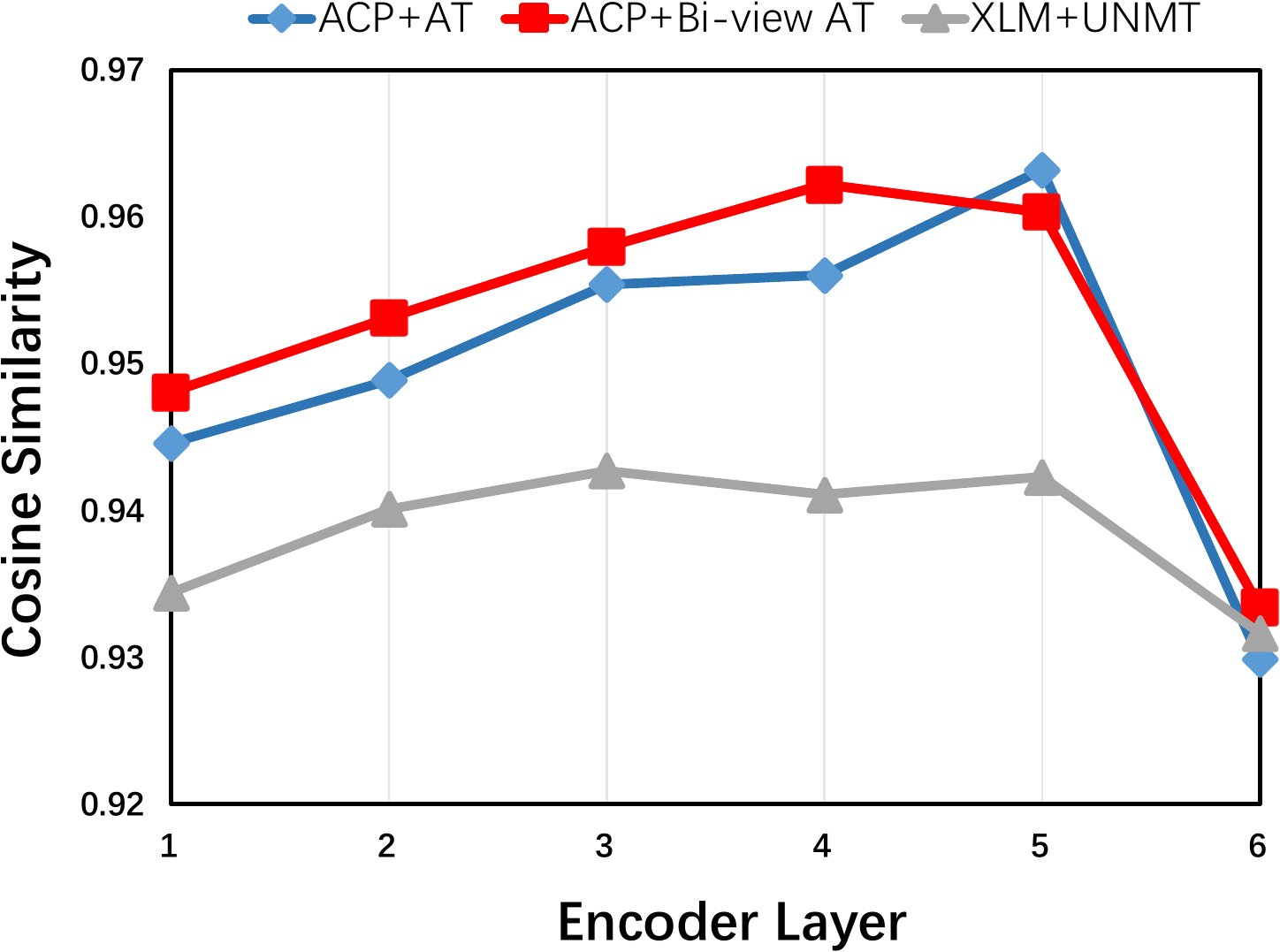} }
\caption{Sentence level cosine similarity of the parallel sentences on each encoder layer.}
\label{fig:sentence}
\end{figure}

Through the word level analysis, we can see that AT/Bi-view AT leads to more consistent word embedding space shared by both languages, making the translation between both languages easier.

\vspace{5 pt}
\noindent  \textbf{Sentence Level Similarity of Parallel Sentences}
\vspace{5 pt}

\noindent We check the sentence level representational invariance across languages for the cross-lingual pretraining methods. In detail, following Arivazhagan et al. \shortcite{arivazhagan2019missing}, we adopt max-pooling operation to collect the sentence representation of each encoder layer for all Chinese-to-English sentence pairs in the test set. Then we calculate the cosine similarity for each sentence pair and average all cosine scores.

Figure \ref{fig:sentence} shows the sentence level cosine similarity. ACP+Bi-view AT consistently has a higher similarity for parallel sentences than XLM+UNMT on all encoder layers. When compare Bi-view AT and AT, the Bi-view AT is better on more encoder layers. 

We can see that in both word level and sentence level analysis, our AT methods achieve better cross-lingual invariance, significantly reduce the gap between the source language space and the target language space, leading to decreased translation difficulty between both languages.

\vspace{5 pt}
\noindent  \textbf{Ground-Truth Dictionary Vs Artificial Dictionary}
\vspace{5 pt}

\begin{table}[htbp]
\small
\begin{tabular}{l|c|c|c|c}
\bottomrule[1.5pt]
& \multicolumn{2}{c|}{Ground-Truth Dict.} & \multicolumn{2}{c}{Artificial Dict.} \\ \hline
& zh$\rightarrow$en & en$\rightarrow$zh & zh$\rightarrow$en & en$\rightarrow$zh  \\ \hline
AT & \textbf{19.83} & \textbf{9.18} & 16.7 &  6.98 \\ 
Bi-view AT & \textbf{21.16} & \textbf{11.23} & 18.23 & 8.50  \\ \hline
 \toprule[1.5pt]
\end{tabular}
 \caption{BLEU of AT methods using either the ground-truth dictionary or the artificial dictionary.}\label{tbl:ground-artificial}
\end{table}

\begin{table}[htbp]
\small
\centering
\begin{tabular}{l|c}
\bottomrule[1.5pt]
XLM+UNMT & 20.68 \\  \hline
ACP+AT with 1/4 of the dictionary & 22.84  \\ 
ACP+AT with 1/2 of the dictionary & 24.32  \\ 
ACP+AT with the full dictionary & 26.80  \\ 
 \toprule[1.5pt]
\end{tabular}
 \caption{BLEU of ACP+AT using different size of the dictionary in zh$\rightarrow$en translation.}\label{tbl:dictionary_size}
\end{table}

\noindent Table \ref{tbl:ground-artificial} presents the comparison in English-Chinese. The ground-truth dictionary is from the Muse dictionary deposit, and the artificial dictionary is generated by unsupervised BLI ~\cite{conneau2017word}. We extract top-$n$ word pairs as the artificial dictionary, where $n$ is the same as the number of entries in the ground-truth dictionary.

Both dictionaries use AT methods for translation. As shown in Table \ref{tbl:ground-artificial}, the ground-truth dictionary performs significantly better than the artificial dictionary in both methods and both translation directions.

\vspace{5 pt}
\noindent  \textbf{The Effect of The Dictionary Size}
\vspace{5 pt}

\noindent We randomly select a portion of the ground-truth bilingual dictionary to study the effect of the dictionary size on the performance. Table \ref{tbl:dictionary_size} reports the performances of ACP+AT using a quarter or a half of the zh$\rightarrow$en dictionary. 

It shows that, in comparison to the baseline of XLM+UNMT that does not use a dictionary, a quarter of the dictionary consisting of around 3k word pairs is capable of improving the performance significantly. More word pairs in the dictionary lead to better translation results, suggesting that expanding the size of the current Muse dictionary via collecting various dictionaries built by human experts may improve the translation performance further. 

\section{Discussion and Future Work}

In the literature of unsupervised MT that only uses non-parallel corpora, Unsupervised SMT (USMT) and Unsupervised NMT (UNMT) are complementary to each other.  Combining them (USMT+UNMT) achieves significant improvement over the individual system, and performs comparable to XLM+UNMT ~\cite{Lample2018Phrase,artetxe2019acl-effective}.

We have set XLM+UNMT as a stronger baseline, and our ACP+AT/Bi-view AT surpasses it significantly. By referring to the literature of unsupervised MT, we can opt to combine ACP+AT/Bi-view AT with SMT. We leave it as a future work.

\section{Conclusion}

In this paper, we explore how much potential an MT system can achieve when only using a bilingual dictionary and large-scale monolingual corpora. This task simulates people acquiring translation ability via looking up the dictionary and depending on no parallel sentence examples. We propose to tackle the task by injecting the bilingual dictionary into MT via anchored training that drives both language spaces closer so that the translation becomes easier. Experiments show that, on both close language pairs and distant language pairs, our proposed approach effectively reduces the gap between the source language space and the target language space, leading to significant improvement of translation quality over the MT approaches that do not use the dictionary and the approaches that use the dictionary to supervise the cross-lingual word embedding transformation. 

\section*{Acknowledgments}

The authors would like to thank the anonymous reviewers for the helpful comments. This work was supported by National Natural Science Foundation of China (Grant No. 61525205, 61673289), National Key R\&D Program of China (Grant No. 2016YFE0132100), and was also partially supported by the joint research project of Alibaba and Soochow University.

\bibliography{acl2020}
\bibliographystyle{acl_natbib}

\end{document}